\documentclass{article}

\usepackage{arxiv}

\usepackage[utf8]{inputenc} 
\usepackage[T1]{fontenc}    
\usepackage{booktabs}       
\usepackage{amsfonts}       
\usepackage{nicefrac}       
\usepackage{microtype}      
\usepackage{graphicx}
\usepackage[numbers]{natbib}
\usepackage{doi}
\usepackage{algorithm, algorithmic}
\usepackage{caption, subcaption}
\usepackage{amsmath,amssymb,amsfonts}

\title{DeepHeteroIoT: Deep Local and Global Learning over Heterogeneous IoT Sensor Data}

\date{} 					

\author{
    Muhammad Sakib Khan Inan \\ Australian Catholic University \\ North Sydney, NSW, Australia \\ \texttt{muhammadsakibkhan.inan@myacu.edu.au}
    \And
    Kewen Liao \\ Australian Catholic University \\ North Sydney, NSW, Australia \\ \texttt{kewen.liao@acu.edu.au}
    \And
    Haifeng Shen \\ Australian Catholic University \\ North Sydney, NSW, Australia \\ \texttt{haifeng.shen@acu.edu.au}
    \And
    Prem Prakash Jayaraman \\ Swinburne University of Technology \\ Hawthorn, VIC, Australia \\ \texttt{pjayaraman@swin.edu.au}
    \And
    Dimitrios Georgakopoulos \\ Swinburne University of Technology \\ Hawthorn, VIC, Australia \\ \texttt{dgeorgakopoulos@swin.edu.au}
    \And
    Ming Jian Tang \\ Atlassian \\ Sydney, NSW, Australia \\ \texttt{mj2tang@gmail.com}
}

\renewcommand{\undertitle}{Accepted for Publication and Presented in EAI MobiQuitous 2023 - 20th EAI International Conference on Mobile and Ubiquitous Systems: Computing, Networking and Services}
\renewcommand{\shorttitle}{Deep Local and Global Learning over Heterogeneous IoT Data}

\begin{document}
\maketitle

\begin{abstract}
Internet of Things (IoT) sensor data or readings evince variations in timestamp range, sampling frequency, geographical location, unit of measurement, etc. Such presented sequence data heterogeneity makes it difficult for traditional time series classification algorithms to perform well. Therefore, addressing the heterogeneity challenge demands learning not only the sub-patterns (local features) but also the overall pattern (global feature). To address the challenge of classifying heterogeneous IoT sensor data (e.g., categorizing sensor data types like temperature and humidity), we propose a novel deep learning model that incorporates both Convolutional Neural Network and Bi-directional Gated Recurrent Unit to learn local and global features respectively, in an end-to-end manner. Through rigorous experimentation on heterogeneous IoT sensor datasets, we validate the effectiveness of our proposed model, which outperforms recent state-of-the-art classification methods as well as several machine learning and deep learning baselines. In particular, the model achieves an average absolute improvement of 3.37\% in Accuracy and 2.85\% in F1-Score across datasets.
\end{abstract}

\keywords{IoT Sensor Data  \and Data Heterogeneity \and Deep Learning \and Time Series Classification}

\section{Introduction}
IoT sensor data\footnote{Aligning with the works of \cite{mace_paper,wise_paper,vculic2023lost}, we define IoT sensor data as sequences of raw numerical readings/observations obtained from IoT sensors over time.} or readings exhibit time series-like sequences. The classification of such sequence data has emerged as an in-demand research area, given the ubiquitous presence of IoT sensors in our daily lives in accordance with the growing dominance of mobile computing \cite{mohammadi2018deep,Senshamart,zhang2022fedgroup}. The next decade is predicted to see IoT devices creating \$14.4 trillion worth of value across different industries \cite{ciscoReport}. IoT sensor devices are widely adopted in the sectors like healthcare, environmental monitoring, traffic management, and energy management, in a diverse range of settings \cite{mace_paper,wise_paper,vculic2023lost}. This naturally causes the IoT data to vary in timestamp range, observation configurations, sampling frequency, unit of measurement, and geo-location \cite{teh2020sensor}, which leads to the production of a plethora of IoT sensor data with inherent heterogeneity \cite{georgakopoulos2016internet,mace_paper,borges2022classification}. The inherent heterogeneity, where sensors of the same type (e.g., temperature) record data in different units of measurement (Celsius or Fahrenheit) or with varying sampling frequencies, results in inconsistent and non-repetitive complex discriminatory patterns in time-series readings from IoT sensors. In such scenarios, time-series sensor readings/sequences display a mixture of local sub-patterns occurring at irregular intervals, forming a global pattern of heterogeneous nature at times. 

The growing use of IoT sensors, integrated with a variety of mobile computing devices, highlights the importance of accurate classification, such as distinguishing between temperature and humidity data, to ensure the reliability of dynamic systems like smart homes~\cite{elazhary2019internet}. The heterogeneous nature of IoT sensor data also affects the data quality \cite{Senshamart}, making the classification tasks challenging even for the existing machine-learning techniques. It also hinders the effective re-purposing of these data for interoperability~\cite{mohammadi2018deep}. For instance, the traditional time series classification methods often underperform~\cite{bagnall2017great,wise_paper} when it comes to classifying IoT sensor data in a multi-class setting \cite{bagnall2017great,borges2022classification}.

Recent research has witnessed the development of machine learning (ML) algorithms that leverage textual metadata (e.g., sensor names and descriptions) to ameliorate the classification of IoT sensor data \cite{mace_paper,wise_paper,Madithiyagasthenna2020}. However, such textual metadata are often not available or presented inaccurately in IoT sensor datasets \cite{borges2022classification}. The expansion of IoT systems also increases the chance of significant metadata loss due to disrupted or corrupted connections to configuration metadata, making collected IoT sensor data unusable \cite{vculic2023lost}. Moreover, depending on the availability of accurate metadata limits the capability and feasibility of learning where only numeric sensor readings are available \cite{borges2022classification,calbimonte2012deriving}. Though optimized ensemble machine-learning algorithms \cite{mace_paper,wise_paper} and feature transformation-based strategies \cite{borges2022classification} recently achieved notable successes in heterogeneous IoT sensor data classification, those approaches lack generalizability and scalability and they are also not deeply learned end-to-end processes \cite{ismail2019deep}. Along with the rapid advancement of artificial intelligence, deep learning (DL) algorithms have demonstrated massive potential in time series classification \cite{ismail2019deep} due to their advantage of representation learning. Some studies \cite{rahman2022intelligent,dua2023inception} developed DL-based solutions for IoT sensor data classification but only considered a homogeneous data domain. There also have been studies that incorporated deep learning-based strategies for IoT sensor data forecasting \cite{pan2020water,sajjad2020novel}. However, there were no DL models specifically designed for heterogeneous IoT sensor data classification. In search of such a DL solution, we first rigorously investigated the state-of-the-art deep time series classification algorithms including InceptionTime \cite{ismail2020inceptiontime} and TapNet \cite{zhang2020tapnet}, along with other traditional machine learning and deep learning based algorithms. From our empirical study, we found that in terms of the heterogeneous IoT sensor datasets, existing solutions do not perform well which was similarly highlighted in previous studies \cite{bagnall2017great,borges2022classification}. Though there exist effective ensemble learning models like MACE \cite{mace_paper} and TKSE \cite{wise_paper}, these models are highly complex, easily overfit, and do not learn the underlying artifacts of heterogeneous IoT sensors with a contextual-learned approach.

To this end, we propose a novel end-to-end deep learning model with the aim of better capturing the heterogeneous patterns of IoT sensor data that would lead to enhanced classification performance. The model learns ensembled \textit{local} neighborhood-based features with Convolutional Neural Network (CNN) convolution layers of varying kernel sizes, and \textit{global} whole series sequential features with a stack of Bi-directional Gated Recurrent Units (GRUs). We deliberately name the model as DeepHeteroIoT for ease of reference.
The main contributions of the work are summarized as follows:
\begin{itemize}
    \item We propose the first end-to-end DL model for heterogeneous IoT sensor data classification. The model facilitates a novel combination of CNN and Bi-directional GRU modules to respectively learn local and global patterns of heterogeneous IoT sensor data.
    \item We provide rigorous empirical comparisons across multiple IoT datasets along with the development of a new heterogeneous IoT sensor dataset. Experimental results demonstrate that our deep learning model achieves state-of-the-art classification performance in both accuracy and F1-score.
\end{itemize}

\section{Related Works}
\paragraph{IoT Sensor Data Classification}
IoT sensors typically generate numerical continuous data, forming numerical sequences when collected at regular or irregular intervals over a specific time period. The task of classifying these continuous numerical series is akin to problems such as Time Series Classification (TSC) or Sequence Classification. To deal with challenging heterogeneous IoT sensor data, Montori et al.~\cite{wise_paper} proposed a two-layer sequential ensemble approach of classifiers where individually trained classifiers are stacked during the prediction pipeline in a sequential manner to filter out classes with fewer prediction probabilities. In their extended version \cite{mace_paper}, an improved multi-layer sequential ensemble model was proposed with novel heuristics for selecting and ordering classifiers and filtering classes between classifiers. Borges et al.~\cite{borges2022classification} proposed a transformation-based classification strategy that converts raw sensor data to an ordinal pattern with improved feature representation and class separability. Postol et al.~\cite{postol2019time} proposed a topological data analysis-based strategy for noisy IoT sensor data classification. A random forest-based strategy was recently proposed to classify a signal type being read from an IoT sensor utilizing raw IoT sequence data \cite{vculic2023lost}. Moreover, a probabilistic data mining approach \cite{calbimonte2012deriving} incorporating slope distribution computation via linear approximation of time series sequence was first developed for heterogeneous IoT sensor data classification. Although previous approaches achieved notable classification performance, they have only sought global patterns from different perspectives. Several studies have explored the potential of deep learning for classifying IoT sensor data for tasks such as intelligent waste management \cite{rahman2022intelligent} and human activity recognition \cite{dua2023inception}. However, the sensors used in these domains are predominantly homogeneous and share similar characteristics. To the best of our knowledge, the development of end-to-end deep learning algorithms to better capture rich low-dimensional temporal semantics of heterogeneous IoT sensor data (that comes from diverse environments) is still missing in the literature.

\paragraph{Time Series Classification}
The time series classification domain has seen many developments over the years such as the classical methods of Dynamic Time Warping (DTW) and Weighted Dynamic Time Warping \cite{bagnall2017great} that rely on whole-series distance-based similarity measures. Additionally, methods that work with the approach of subsequence extraction, coined as "shapelets" based strategies, have been popular in data mining \cite{ye2009timeShapelets,rakthanmanon2013fastShapelets}. However, to deal with the dimensionality issue present in the distance measure-based approaches, a lower-bound symbolic representation of time series sequence named Symbolic Aggregation approXimation (SAX) \cite{lin2007experiencing} was introduced to first transform data via Piecewise Aggregate Approximation (PAA) and then symbolize the PAA representation into a discrete string. Lin et al.~\cite{lin2012rotation} proposed a histogram-based high-level feature extraction strategy utilizing PAA through SAX representation for time series classification, taking inspiration from the popular "bag of words" strategy in the domain of text mining.  Nevertheless, the majority of conventional techniques concentrate on individual time series attributes like shape or frequency. These methods lack scalability and efficacy in handling the diverse characteristics inherent to the IoT data \cite{mace_paper}. In recent years, learning-based methods that incorporate state-of-the-art deep learning algorithms have been developed for time series classification due to their capability of learning low-dimensional feature representations automatically without domain knowledge \cite{ismail2019deep}. Convolutional Neural Network (CNN) based model InceptionTime \cite{ismail2020inceptiontime} and Attention-mechanism incorporated hybrid approach TapNet \cite{zhang2020tapnet} are two prominent examples of such effective deep learning-based solutions for time series classification. Additionally, ensemble or hybridization of deep learning algorithms including CNN and recurrent neural networks (RNN) have shown great potential in extracting rich semantics of time series in terms of sensor data classification \cite{interdonato2019duplo} and forecasting challenges in various domains such as water level prediction \cite{pan2020water} and short-term residential load forecasting \cite{sajjad2020novel}. While current state-of-the-art time series classification algorithms exhibit strong performance on general time series data, they often lack the specialized design required to effectively capture the inherent heterogeneity within IoT sequence data. As a consequence, there is a critical need to develop novel methodologies that can accommodate the varying characteristics of IoT sequence data and extract meaningful insights from these distinct and heterogeneous sources.


\section{DeepHeteroIoT: Our Proposed Deep Learning Model}
\begin{figure}[h!]
    \centering
\includegraphics[scale=0.7,width=0.95\textwidth]{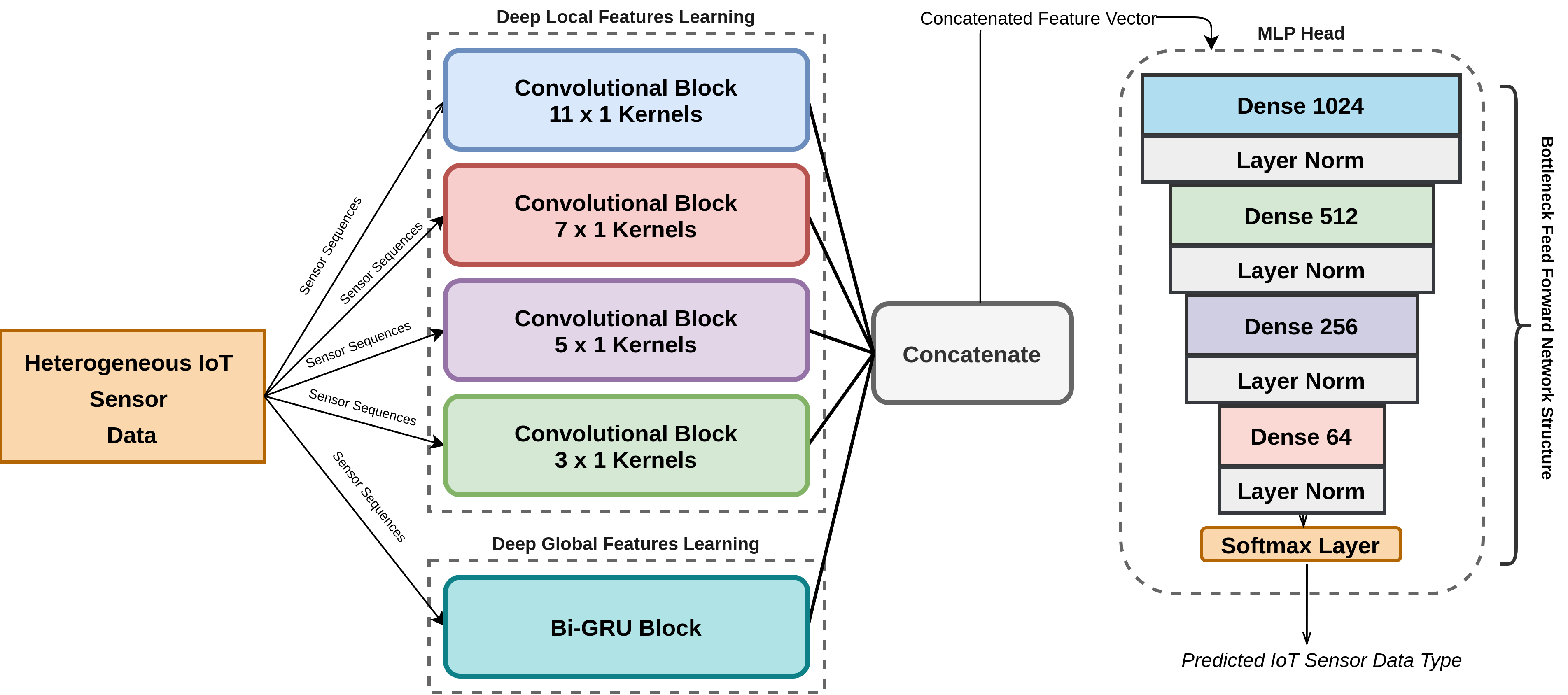}
    \caption{An overview of our proposed deep learning model with  learnable local and global learnable patterns for heterogeneous IoT sequence data classification}
    \label{fig:proposed-framework}
\end{figure}

In this section, we delineate a brief description of the key components of our proposed end-to-end deep learning model. As represented in Figure \ref{fig:proposed-framework}, our proposed deep learning model, DeepHeteroIoT, extracts learned feature representation of global patterns using a stack of bi-directional GRU and local patterns using an ensemble of decoupled CNN stacks with varying kernel sizes. Learned local and global feature vectors are then combined through a concatenation layer which connects to a multilayer perceptron (MLP) head for final classification.

\paragraph{\textbf{Deep Local Features.}}
Initially developed to tackle complex issues within Computer Vision, Convolutional Neural Networks (CNNs) leverage local neighborhood-based features to enhance the distinction between different classes by shaping decision boundaries. As the realm of deep learning continues its rapid evolution, CNN architectures have showcased significant potential in extracting robust spatial features from input sequences. By employing convolutional operations with small kernel sizes, CNNs excel at capturing local intricacies, making them an excellent fit for intricate time series or sequence analyses. In our proposed deep learning model, we introduce a customized novel CNN architecture tailored to seamlessly integrate into our end-to-end framework for IoT sensor data classification. A single convolutional block incorporated in our proposed model can be expressed mathematically;
\begin{equation}\label{eq:conv}
F_{f}=ConvBlock(X,f,p)
\end{equation}

In equation \ref{eq:conv}, $X$ represents a vector containing input sequences such as [$X_0$, $X_1$, ... ... , $X_t$], distributed over $t$ timestamps, $f$ indicates kernel size for convolutional layers, $p$ indicates the type of padding to be incorporated and $F_f$ is the feature space computed base on kernel size $f$. Also, each $ConvBlock$ consists of a stack of convolutional layers with designated kernel size and pooling layers. Each set of convolutional blocks comprises nine consecutive 1D-convolutional layers. Following the 3rd and 6th convolutional layers, a max-pooling layer with a size of 2 is incorporated. Each convolutional block ends with a Global Average Pooling layer \cite{global_average_pooling}. The pooling operations allow the models to downsample feature maps in the way the most significant information has been retained and passed to the next layer while reducing the spatial dimensions. 
Within each convolutional block, the sequence of stacked convolutional layers is set up with 128 filters for the layers preceding the initial max-pooling stage, and subsequently, 64 filters are employed for all successive convolutional layers.

The size of the kernels used in convolutional layers is of utmost importance for capturing local features, as they function as windows to compute features on sub-patterns extracted from the entire time series like sequences at a lower level. In algorithm \ref{algo:framework}, local deep learned features are computed at steps 2,3,4 and 5 incorporating $X$ into equation \ref{eq:conv} for varying kernel sizes with designated padding strategy. Drawing inspiration from the idea of incorporating kernels (convolutional windows) of different sizes to enhance the extraction of local features from different spatial ranges \cite{szegedy2016rethinking}, we propose a decoupled ensemble structure for our convolutional module. In our design, we stack convolutional layers with kernel sizes of 3, 5, 7, and 11, allowing our model to learn sub-patterns using various receptive field sizes in separate blocks. These small kernel sizes were adopted on the basis of their proven computational efficiency in prior state-of-the-art CNN models \cite{szegedy2016rethinking}, as well as their effectiveness in capturing local features in complex computer vision problems. The final choice of kernel sizes was also determined from randomized parameter tuning. Additionally, padding plays a critical role in the design of CNNs as it ensures the preservation of spatial dimensions in input sequences. By retaining spatial information, padding enables CNNs to process time series alike sequences with varying lengths consistently. For the convolutional layers, we use 'causal' padding. This type of padding strategy adds zeros at the start of the input sequence after each convolution operation. Importantly, this maintains the temporal sequence order, ensuring that $output_t$ does not depend on $input_{t+1}$ to $input_{t+n}$ \cite{vanwavenet}.

\paragraph{\textbf{Deep Global Features.}}
Gated Recurrent Unit (GRU), one type of Recurrent Neural Network (RNN), has emerged as a powerful deep learning technique for capturing overall long temporal dependencies or global patterns in sequential data for challenges like time series classification \cite{yamak2019comparison,Elsayed2019}. Though GRU possesses a simpler architecture with fewer gates and parameters to train,  it still excels at capturing both short-term and long-term dependencies in sequential data compared to Long Short-Term Memory (LSTM) \cite{yamak2019comparison}. These unique traits make it computationally more efficient and potentially better suited for our IoT Sensor data classification challenge. A GRU unit maintains a gating mechanism incorporating an Update Gate ($u_{t}$) and Reset Gate ($r_{t}$) based strategy to selectively update and reset its internal state for capturing long-term dependencies and global facets or features of sequence data along with solving the vanishing gradient problem. 
Mathematically,

\begin{align}
        u_{t}=\sigma(W_{u} \odot  x_{t}+U_{z} \odot  h_{t-1}+b_{u})\\
        r_{t}=\sigma(W_{r} \odot x_{t}+U_{r} \odot h_{t-1}+b_{r})
\end{align}

Here, $t$ denotes the timestamp, $x_{t}$, and $h_{t-1}$ accordingly denotes the input in the current timestamp and hidden state in the previous timestamp. $W$, $U$ and $b$ are respective weight matrices and vectors for update and reset gates. To better capture the global sequential patterns of IoT sensor data, we develop a stacked Bi-directional GRU module for integration into our deep learning model. Bi-directional means capturing the patterns of the sequence data from both directions (forward and backward) to allow it to learn temporal dependencies and patterns that exist in both past and future contexts. A single block of bi-directional GRU unit, incorporated into our deep learning model can be represented mathematically as;
\begin{equation}\label{eq:gru}
gru_{i}=Bidirectional(GRU(X, D))
\end{equation}
Here, $X$ represents a vector containing input sequences such as [$X_0, X_1, ... ..., X_t$], distributed over $t$ timestamps, and $D$ represents dimensions of output representation space computed by the $GRU$ function. $gru_{i}$ is the output sequence returned by one $Bidircetional$ function that concatenates the computed output features by the $GRU$ function from both the forward and backward direction in an appending manner. As described in Figure \ref{fig:proposed-framework} and Algorithm \ref{algo:framework}, for deep global sequential feature extraction we stack 3 Bi-directional GRU units accompanied by Batch Normalization after each unit. The hyperparameter setting of parameter $D$ for each unit is accordingly 128, 128, and 64 in steps 7, 8, and 9 for algorithm \ref{algo:framework} with the incorporation of equation \ref{eq:gru} for computation of global patterns at each epoch for every sample in $X$.

\paragraph{\textbf{Combined Deep Learning Model.}}
To improve the classification performance on challenging IoT data, as delineated in Algorithm \ref{algo:framework}, we propose a novel deep learning model that incorporates deep learning-based local features by our customized CNN and global features by our Bi-directional GRU. As depicted in Algorithm \ref{algo:framework}, the learned feature space at every epoch from both of these networks is concatenated through a concatenation layer in step 10 (Algo. \ref{algo:framework}) and fed into an MLP Head (Multi-layer perception-based fully-connected layer) in step 11 (Algo. \ref{algo:framework})  which consists of a stack of dense feed-forward layers accompanied by layer-normalization after each dense linear projection layer. The stack of dense linear projection layers follows a bottleneck design (accordingly 1024, 512, 256 and 64) in terms of the number of neurons in each layer inspired by the previous state-of-the-art architectures \cite{howard2017mobilenets}. The key motivation for this type of design is to reduce the number of parameters in subsequent layers to force the network to learn more compact representations that eventually improve generalization reducing the chance of overfitting \cite{howard2017mobilenets}. The number of $Epochs$ was set to 200 using ADAM (Adaptive Moment Estimation) optimizer with a learning rate of 0.001. The weights of the best epoch on the validation stage were considered for the best model.
\begin{algorithm}[h!]
\caption{DeepHeteroIoT: Our Proposed Deep Learning Model}
	\label{algo:framework}
    	\hspace*{\algorithmicindent}{\textbf{Input:} $\mathbb{X}$: 2-D matrix that contains $N$ number of individual sequences represented as $x_i$ where IoT Sensor data sequences is distributed of $t$ time-stamps. $\mathbb{X}$ = [$x_i$, $x_{i+1}$, ... , $x_{i+N}$]. Each $x_i$ contains a sequence of data points representing IoT sensor readings.}. Every $x_i$ in $\mathbb{X}$ feeds as input to the neural network over the epochs into individual steps. 
	\begin{algorithmic}[1]
	\FOR{Each $epoch$ in $Epochs$} 
		\STATE $F_{3}$ = $ConvBlock(X, 3, "causal")$
            \STATE $F_{5}$ = $ConvBlock(X, 5, "causal")$
            \STATE $F_{7}$ = $ConvBlock(X, 7, "causal")$
            \STATE $F_{11}$ = $ConvBlock(X, 11, "causal")$
            \STATE Concatenate $F_{3}$, $F_{5}$, $F_{7}$, $F_{11}$ in an appending manner which generates a vector         $concat_{local}$, containing a matrix of local features.
            \STATE $gru_1$ = $Bidirectional(GRU(X, 128))$
            \STATE $gru_2$ = $Bidirectional(GRU(gru_1, 64))$
            \STATE $gru_3$ = $Bidirectional(GRU(gru_2, 64))$
            \STATE Concatenate extracted local feature ($concat_{local}$) and global features ($gru_3$) in an appending manner.
            \STATE Feed the concatenated feature space into the MLP Head.
		\STATE Compute softmax predictions.
            \STATE Compute loss of training epoch.
            \STATE Update weights of each layer based on the learning rate.
	\ENDFOR
	\end{algorithmic}
	\hspace*{\algorithmicindent}{\textbf{Output:} $\mathbb{O}$: predicted class labels defining IoT sensor data }
\end{algorithm}


\section{Experiments}
In this section, we outline the outcomes of our model's experiments using benchmark IoT datasets\cite{mace_paper,wise_paper}. To evaluate the efficacy of our novel deep learning model, we contrasted its Accuracy and F1 scores with conventional machine learning and deep learning models. We also compared it to time series classification-centric deep learning models like InceptionTime \cite{ismail2020inceptiontime} and TapNet \cite{zhang2020tapnet}. Furthermore, we conducted a comparison between our model and a contemporary state-of-the-art ensemble machine learning model known as MACE (as presented in \cite{mace_paper}), which was explicitly tailored for classifying IoT data in a prior research endeavor. We omitted comparisons with feature-transformation or data-mining oriented methodologies \cite{borges2022classification} for classifying IoT sensor data. Because, these techniques rely on statistical transformations of IoT sequence data and our study emphasizes a comprehensive exploration of the capabilities of ML and DL methods in capturing heterogeneous patterns within IoT sequence data at a more granular level \cite{ismail2019deep}. To ensure a consistent comparison within our domain and alignment with the methodology in the prior study \cite{mace_paper}, we adopted the identical experimental design, including the division of datasets into training and testing subsets. We performed a stratified split over
the datasets assigning 70 \% to the training set and 30 \% to the testing using the Scikit-Learn library by setting parameter value "random\_state=100". The tested classifiers in this paper are implemented in Python 3.9.16 on top of Scikit-Learn for machine learning models (except XGBoost) and Tensorflow for deep learning models. For XGBoost, we utilize the Python implementation of the XGBoost algorithm. All the experiments were conducted on a Linux server running Ubuntu 20.04 LTS as the OS, equipped with a 24-core Intel CPU (x86\_64) and 128 GiB of RAM. In our work, we include the Swiss Experiment dataset and Urban Observatory dataset from MACE \cite{mace_paper} paper. We exclude the ThingSpeak dataset which was included in the MACE \cite{mace_paper} paper as this dataset contains textual meta-data describing IoT sensors which do not go with the motivation of this study of dealing with IoT sensor classification challenge utilizing only raw numeric sensor observations. Additionally, to validate the generalizability of our proposed solution,  we constructed another dataset named IOWA ASOS, inspired by a recent study \cite{borges2022classification} which will be discussed in detail in the remaining sections. 

\subsection{IoT Sensor Datasets}
In this section, we delineate the three benchmark IoT Sensor datasets that have been utilized in this study including the IOWA ASOS dataset that we have developed specifically for this paper to prove the generalizability of our model performance along with Urban Observatory (UrbObs) and Swiss Experiment (Swiss) datasets from previous studies \cite{mace_paper,calbimonte2012deriving}.
\begin{itemize}
\item \textbf{IOWA ASOS (IOWA).} Inspired by the study of \cite{borges2022classification}, we developed this IoT sensor dataset by collecting sensor data of various meteorological sensors from the data archive of Iowa Environmental Mesonet (IEM) prepared by the IOWA State University and made available via API and web interface \footnote{Iowa Environmental Mesonet ASOS-METAR Data Download: https://mesonet.agron.iastate.edu/ASOS/}. This dataset contains sensor data from meteorological sensors at airports using Automated Surface Observing Systems (ASOS) \footnote{Automated Surface Observing Systems from the U.S. National Weather Service: https://www.weather.gov/asos/asostech.}. These sensors can produce observations every minute or every hour based on the requirements of the airport authority to corroborate aviation operations by using weather forecasting reports. To make our own version of the new IoT dataset for this study, we collect data with 1-hour intervals over a 6-month time-span from 5 different stations in the IOWA state of the United States. 
To make this dataset suitable for an IoT Sensor data classification challenge, we include 1-week data at 1-hour intervals for each individual classification sample. The dataset contains 8 class labels and these are: Air Temperature, Dew Point Temperature, Relative Humidity, Wind Direction, Pressure Altimeter, Visibility, Wind Gust, and Apparent Temperature (Heat Index).  The missing values are imputed by the standard imputation strategy \cite{borges2022classification} of replacing by global average (mean) on that time-stamp across the individual class labels.

\item \textbf{Urban Observatory (UrbObs).} The version of this dataset we utilized for our experimentation was developed in a previous study \cite{mace_paper} of IoT sensor data classification. We have incorporated the exact same version of this dataset to make our experimental results comparable to the literature. This dataset contains sensor data generated through an innovative initiative \footnote{Urban Observatory Environment Program: http://newcastle.
urbanobservatory.ac.uk/} spearheaded by the University of Newcastle. The program aims to establish an urban sensor network in Newcastle, United Kingdom (UK) and provides publicly accessible real-time environmental data \cite{james2014urban}. The dataset has 16 different class labels including $NO_{2}$ (Nitrogen Dioxide), Wind Direction, Humidity, Wind Speed, Temperature, Pressure, Wind Gust, Rainfall, Soil Moisture, Average Speed, Congestion, Traffic Flow, Journey time, Sound, CO (Carbon Monoxide) and NO (Nitrogen Monoxide). This dataset contains highly correlated IoT data, primarily due to its city-wide scope and originating from a single source \cite{mace_paper}.
\item \textbf{Swiss Experiment (Swiss).} This dataset is one of the few extremely noisy and heterogeneous IoT datasets that consist of data from various microscopic locations in various time-stamps within the Swiss Alps Mountain range \cite{calbimonte2012deriving}. This dataset contains data from sensors having a diverse range of sampling rates making the phase-shift of data series very significant \cite{mace_paper,calbimonte2012deriving}. Following the exact same version of this dataset from the previous study \cite{mace_paper}, we cut each
time series data to the length of the shortest stream as the original data contains data of slightly different lengths. It includes class labels such as $CO_{2}$ (Carbon Dioxide), Humidity, Lysimeter, Moisture, Pressure, Radiation, Snow Height, Temperature, Voltage, Wind Speed, and Wind Direction. 

\end{itemize}
\begin{table}[h]
\centering
\caption{Summary of IoT Datasets}
\label{tab:datasets-summary}
\begin{tabular}{|l|r|r|r|r|}
\hline
\textbf{Dataset} &
  \multicolumn{1}{c|}{\textbf{Length}} &
  \multicolumn{1}{c|}{\textbf{Duration}} &
  \multicolumn{1}{c|}{\textbf{Samples}} &
  \multicolumn{1}{c|}{\textbf{Labels}} \\ \hline
Urban Observatory & 864 & 1 day    & 1065 & 16 \\ \hline
Swiss Experiment  & 445 & Variable & 346  & 11 \\ \hline
IOWA ASOS         & 168 & 1 week   & 1000 & 8  \\ \hline
\end{tabular}
\end{table}

Table \ref{tab:datasets-summary} provides a summary of 3 IoT datasets highlighting time series sequence length, duration of record per sample, the total number of samples, and the total number of class labels for each dataset. Out of these three IoT datasets, the Swiss Experiment dataset exhibits a notable class imbalance and has a very limited number of training samples, making it a particularly challenging dataset for deep learning methods. On the other hand, the IOWA ASOS dataset is almost completely balanced where each class has almost the same amount of data samples. The Urban Observatory exhibits some degree of class imbalance, but due to the presence of less noisy IoT sensor data, the imbalance issue does not pose a major problem for this dataset\cite{mace_paper}. In the Swiss Experiment, the majority class has 78 samples, while the minority class has only 14 samples in total.

\subsection{Experimental Results}
In this section, a detailed representation of experimental results along with a rigorous comparison against baseline machine learning and deep learning models have been delineated. Also, performance comparison with previous state-of-the-art models across 3 datasets is presented and validated.
\paragraph{\textbf{Evaluation Metrics.}}
Before starting, we describe the evaluation metrics that have been computed and compared throughout the experimental validation stage for our proposed model and all other experiments. These are:
\begin{itemize}
    \item \textit{Accuracy:} It indicates the percentage of IoT Sensor data that have been classified correctly by our model out of total data samples. 
    \item \textit{F1-Score:} It is a harmonic mean of Precision and Recall which gives us a balanced perception of our model's performance and helps to validate DL model's effectiveness properly. 
\end{itemize}
For the evaluation of each experiment, we focus on the Accuracy and weighted average F1-Score throughout the study. Later, we computed the macro-average F1 Score for comparison of the model's effectiveness with a previous state-of-the-art study \cite{mace_paper}. The weighted average F1 Score is a more suitable way to compute F1-Scores for scenarios where the class imbalance is present to find the dataset. On the other, the macro average F1-Score does not take into account that class imbalance is present in the dataset.

\paragraph{\textbf{Ablation Studies.}} To validate the efficacy of our proposed combination of global and local deep learning-based learned patterns for IoT sequence data, we illustrate a detailed ablation analysis of our proposed model in Table \ref{tab:ablation-all}  with Accuracy scores and weighted average F-1 scores across three different IoT datasets. From the results of Table \ref{tab:ablation-all}, it is clearly visible that our proposed novel deep-learning model outperforms individual components including only global features, only local features, and only MLP Head (without any deep feature extraction) based classifiers with a high margin. However, out of individual components, classification with only global features dominates compared to combinations with only local features and only MLP Head across 3 datasets.

\begin{table}[h!]
\centering
\caption{Ablation study of our proposed deep learning model across IoT datasets}
\label{tab:ablation-all}
\begin{tabular}{l|r|r|r|r|r|r|}
\cline{2-7}
 &
  \multicolumn{1}{l|}{\begin{tabular}[c]{@{}l@{}}UrbObs \\ (Accuracy)\end{tabular}} &
  \multicolumn{1}{l|}{\begin{tabular}[c]{@{}l@{}}UrbObs \\ (F1-Score)\end{tabular}} &
  \multicolumn{1}{l|}{\begin{tabular}[c]{@{}l@{}}Swiss \\ (Accuracy)\end{tabular}} &
  \multicolumn{1}{l|}{\begin{tabular}[c]{@{}l@{}}Swiss \\ (F1-Score)\end{tabular}} &
  \multicolumn{1}{l|}{\begin{tabular}[c]{@{}l@{}}IOWA \\ (Accuracy)\end{tabular}} &
  \multicolumn{1}{l|}{\begin{tabular}[c]{@{}l@{}}IOWA \\ (F1-Score)\end{tabular}} \\ \hline
\multicolumn{1}{|l|}{Global Features} & 90.63\%          & 89.92\%          & 79.81\%          & 78.5\%             & 94.33\%          & 94.32\%          \\ \hline
\multicolumn{1}{|l|}{Local Features}  & 88.13\%          & 87.72\%          & 77.88\%          & 76.41\%          & 87.33\%          & 87.50\%          \\ \hline
\multicolumn{1}{|l|}{MLP Head}        & 77.51\%          & 76.02\%          & 45.19\%          & 42.04\%          & 86.01\%          & 85.80\%          \\ \hline
\multicolumn{1}{|l|}{\textbf{DeepHeteroIoT}}           & \textbf{94.38\%} & \textbf{94.27\%} & \textbf{86.54\%} & \textbf{85.12\%} & \textbf{96.01\%} & \textbf{95.93\%} \\ \hline
\end{tabular}
\end{table}
It is also noticeable that only MLP Head performs badly, indicating the importance of deep-learning-based feature extraction before training a fully-connected network-based classifier for the classification of heterogeneous IoT sensor data.  If we draw attention to the accuracy scores presented in Table \ref{tab:ablation-all}, we can notice that our proposed deep learning model outperforms the second-best combination in Table \ref{tab:ablation-all} by 3.75\%, 6.73\% and 1.68\% accordingly on Urban Observatory, Swiss Experiment and IOWA ASOS datasets. Similarly, in terms of weighted average F1-score our proposed model outperforms by 4.35\%,   6.62\%, and 1.61\%. The outperforming scores achieved by our proposed model justify the importance of learning both global and local deep learning-based patterns of IoT sequence data within one combined and end-to-end learning model. As IoT sensor data is heterogeneous, the proposed combination of both local and global learnable deep neural network architecture is a more suitable approach to capture different dynamic facets of IoT sensor data. 

\begin{figure*}[h!]
\begin{subfigure}{0.32\textwidth}
  \centering
  \includegraphics[scale=0.27]{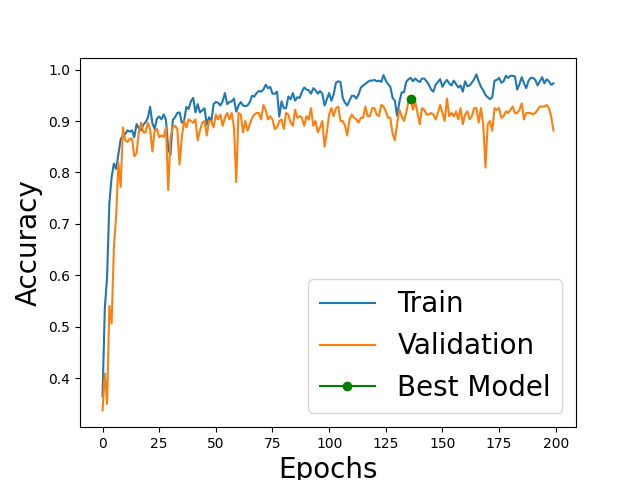}  
  \caption{Urban Observatory}
  \label{subfig:urban-learning-curve}
\end{subfigure}
\begin{subfigure}{0.32\textwidth}
  \centering
  \includegraphics[scale=0.27]{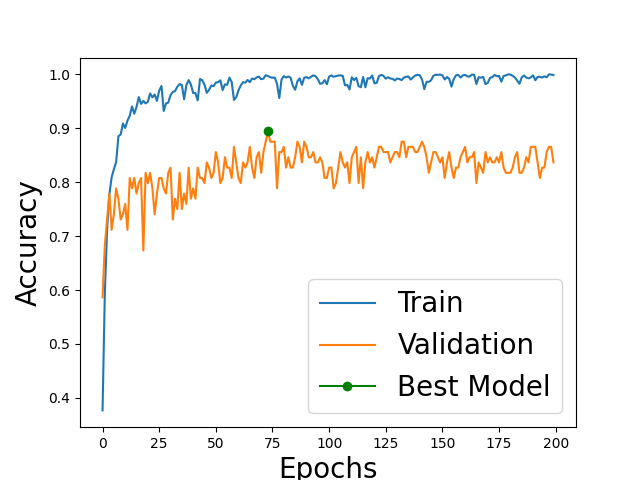}  
  \caption{Swiss Experiment}
  \label{subfig:swiss-learning-curve}
\end{subfigure}
\begin{subfigure}{0.32\textwidth}
  \centering
  \includegraphics[scale=0.27]{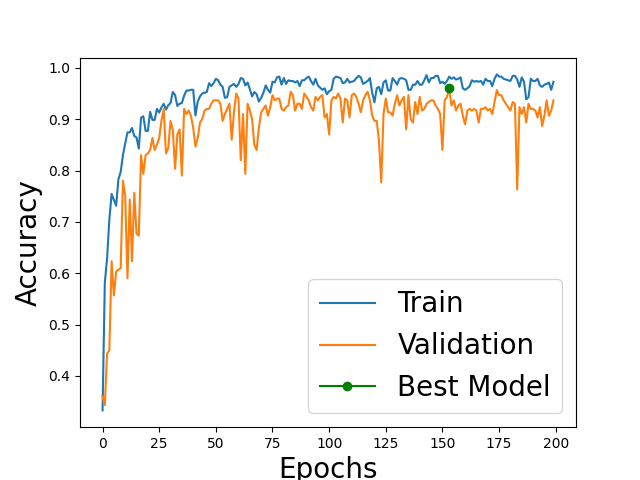}  
  \caption{IOWA ASOS}
  \label{subfig:iowa-learning-curve}
\end{subfigure}
\caption{Learning-curve representation for training and validation (per-epoch) for our proposed model across 3 datasets}
\label{fig:learning-curve-all-datasets}
\end{figure*}

Out of the 3 IoT datasets, as described earlier, the Swiss Experiment dataset is highly heterogeneous with a very limited number of available samples including class imbalance issues. In such scenarios, intricate deep-learning architecture like ours tends to suffer from non-convergence issues and poor training generalizability due to data scarcity issues. For this, only the Swiss Experiment dataset was pre-processed in the training stage by incorporating the time series data-augmentation method followed by oversampling by B-SMOTE \cite{han2005borderline}. Throughout our evaluation process, we also conducted experiments involving the data augmentation module on the other two datasets. However, since these datasets had a sufficient number of samples and exhibited either fewer or no significant class imbalance issues in their original training data, it was obvious that the improvements in accuracy and f1-scores were not as substantial in comparison to the Swiss Experiment dataset after applying data augmentations. And so, for later experimental results only for the Swiss Experiment dataset, we present the results that have been achieved by our model after incorporating the augmented training data. An illustrated representation of the learning curves delineating Accuracy per epoch for the training phase of our proposed deep learning model is outlined in Figure \ref{fig:learning-curve-all-datasets}. The learning curves (Figure \ref{fig:learning-curve-all-datasets}) certainly validate the training efficacy and robust learning capacity of our proposed model across 3 IoT datasets.

\paragraph{\textbf{Comparison of performance against other models.}}
In this stage, we compared our novel DL approach with traditional baseline ML models, including K Nearest Neighbor (KNN), Logistic Regression (LR), Random Forest (RF), Gradient Boosting (GB), Extreme Gradient Boosting (XgBoost), and Support Vector Machine (SVM). To ensure baseline settings for the machine learning algorithms that have been used from Scikit-learn's package, default parameter settings for every algorithm are utilized. For our baseline setting of DL models such baseline RNN and baseline CNN, we accordingly incorporate the Simple RNN API \footnote{ https://www.tensorflow.org/api\_docs/python/tf/keras/layers/SimpleRNN} from Tensorflow with a filter size of 32 and Conv1D API \footnote{https://www.tensorflow.org/api\_docs/python/tf/keras/layers/Conv1D} by Tensorflow and follow the traditional example \footnote{ https://www.tensorflow.org/tutorials/images/cnn} of stacking Convolutional units (3 by 1 kernel sizes and "relu" with no padding) followed by a MaxPooling layer. For time series DL models such as InceptionTime \cite{ismail2020inceptiontime} and TapNet \cite{zhang2020tapnet}, the sktime implementation was utilized \footnote{https://www.sktime.net/en/latest/api\_reference/classification.html}. We trained all the DL models for 200 epochs. 

Table \ref{tab:comp-all}, depicts a rigorous comparison of our DeepHeteroIoT model against baseline ML and DL methods in terms of accuracy and weighted average F1-Score across 3 IoT datasets. Among all machine learning classifiers, only ensemble tree-based models, such as RF, GB, and XgBoost, seem to achieve competitive performance against our proposed deep learning model. Non-tree-based classifiers, such as LR, SVC, and KNN, demonstrate comparatively poor performance across all three datasets. Traditional CNN outperforms traditional RNN in UrbObs and IOWA datasets but lags behind in terms of the Swiss dataset. Notably, the popular InceptionTime model \cite{ismail2020inceptiontime} exhibited very poor performance with a lower accuracy of 29.81\% and an F1-score of 20.15\% on the Swiss dataset. Another such model, TapNet \cite{zhang2020tapnet}, also failed to outperform the tree-based ensemble ML classifiers. The results shed light on the ineffectiveness of traditional time-series classifiers and traditional DL and ML models in heterogeneous IoT sensor scenarios, as discussed earlier in the literature review. Whereas, the potential of ensemble tree-based ML models is worth highlighting for their effective classification in heterogeneous IoT sensor scenarios. From the results, it is apparent that DeepHeteroIoT outperforms every learning-based baseline classifier by a significant margin, that is, achieving a greater average accuracy by absolute 9.29\% and F1-score by 10.07 \% across 3 datasets compared to the second-best performing models. Furthermore, RF emerged as the second-best model for both the Swiss and IOWA datasets, while GB outperformed RF by a slight margin on the UrbObs dataset.
\begin{table}[h!]
\centering
\caption{Comparison of performance against state-of-the-art baseline machine learning and deep learning models across 3 IoT datasets. The best results are in bold and the second best are underlined.}
\label{tab:comp-all}
\begin{tabular}{l|r|r|r|r|r|r|}
\cline{2-7}
 &
  \multicolumn{1}{l|}{\begin{tabular}[c]{@{}l@{}}UrbObs \\ (Accuracy)\end{tabular}} &
  \multicolumn{1}{l|}{\begin{tabular}[c]{@{}l@{}}UrbObs \\ (F1-Score)\end{tabular}} &
  \multicolumn{1}{l|}{\begin{tabular}[c]{@{}l@{}}Swiss \\ (Accuracy)\end{tabular}} &
  \multicolumn{1}{l|}{\begin{tabular}[c]{@{}l@{}}Swiss \\ (F1-Score)\end{tabular}} &
  \multicolumn{1}{l|}{\begin{tabular}[c]{@{}l@{}}IOWA\\ (Accuracy)\end{tabular}} &
  \multicolumn{1}{l|}{\begin{tabular}[c]{@{}l@{}}IOWA \\ (F1-Score)\end{tabular}} \\ \hline
\multicolumn{1}{|l|}{CNN}       & 78.13\%          & 77.71\%          & 47.11\%          & 41.11\%          & 83.01\%          & 82.43\%          \\ \hline
\multicolumn{1}{|l|}{RNN}       & 50.31\%          & 42.87\%          & 58.65\%          & 51.11\%          & 65.01\%          & 62.27\%          \\ \hline
\multicolumn{1}{|l|}{InceptionTime\cite{ismail2020inceptiontime}}       & 64.98\%          & 53.80\%          & 29.81\%          & 20.15\%          & 87.98\%          & 87.99\%          \\ \hline
\multicolumn{1}{|l|}{TapNet\cite{zhang2020tapnet}}       & 53.44\%          & 48.36\%          & 42.31\%          & 33.45\%          & 55.68\%          & 54.52\%          \\ \hline
\multicolumn{1}{|l|}{LR} & 55.63\%          & 51.75\%          & 23.08\%          & 15.52\%          & 53.01\%          & 50.41\%          \\ \hline
\multicolumn{1}{|l|}{KNN}                & 77.19\%          & 73.78\%          & 55.77\%          & 52.36\%          & 75.34\%          & 74.18\%          \\ \hline
\multicolumn{1}{|l|}{SVC}                & 55.63\%          & 47.98\%          & 45.19\%          & 33.85\%          & 55.67\%          & 46..95\%         \\ \hline
\multicolumn{1}{|l|}{RF}      & 86.56\%          & 86.63\%          & \underline{73.07\%}& \underline{70.49\%}& \underline{92.00\%}& \underline{91.84\%}\\ \hline
\multicolumn{1}{|l|}{GB}  & \underline{86.87\%}& \underline{86.74\%}& 65.39\%          & 64.53\%          & 91.33\%          & 91.35\%          \\ \hline
\multicolumn{1}{|l|}{XgBoost}            & 85.63\%          & 85.27\%          & 64.42\%          & 65.36\%          & 90.68\%          & 90.63\%          \\ \hline
\multicolumn{1}{|l|}{\textbf{DeepHeteroIoT}}          & \textbf{94.38\%} & \textbf{94.27\%} & \textbf{89.42\%} & \textbf{89.07\%} & \textbf{96.01\%} & \textbf{95.93\%} \\ \hline
\end{tabular}
\end{table}

\begin{table}[h!]
\centering
\caption{Comparison of performance across 3 IoT datasets with previous state-of-the-art study in IoT sensor data classification domain}
\label{tab:comp-with-mace-all}
\begin{tabular}{l|r|r|r|r|r|r|}
\cline{2-7}
 &
  \multicolumn{1}{c|}{\begin{tabular}[c]{@{}c@{}}UrbObs\\(Accuracy)\end{tabular}} &
  \multicolumn{1}{c|}{\begin{tabular}[c]{@{}c@{}}UrbObs\\ (macro \\ F1-Score)\end{tabular}} &
  \multicolumn{1}{c|}{\begin{tabular}[c]{@{}c@{}}Swiss\\ (Accuracy)\end{tabular}} &
  \multicolumn{1}{c|}{\begin{tabular}[c]{@{}c@{}}Swiss\\ (macro \\ F1-Score)\end{tabular}} &
  \multicolumn{1}{c|}{\begin{tabular}[c]{@{}c@{}}IOWA\\ (Accuracy)\end{tabular}} &
  \multicolumn{1}{c|}{\begin{tabular}[c]{@{}c@{}}IOWA\\ (macro \\ F1-Score)\end{tabular}} \\ \hline
\multicolumn{1}{|l|}{MACE (Top-k)}        & 90.6\%            & 87.9\%           & 84.6\%           & 83.2\%           & 87.33\%          & 87.40\%          \\ \hline
\multicolumn{1}{|l|}{MACE (PF)}           & 90.6\%            & 87.9\%           & 83.2\%           & 83.2\%           & 87.33\%          & 87.40\%          \\ \hline
\multicolumn{1}{|l|}{MACE (SoF)}          & 90.6\%            & 87.9\%           & 82.7\%           & 80.0\%           & 90.33\%          & 90.48\%          \\ \hline
\multicolumn{1}{|l|}{MACE (brute-force)} & 92.2 \%           & 88.7\%           & 86.5\%           & 84.0\%           & 91.01\%          & 91.04\%          \\ \hline
\multicolumn{1}{|l|}{\textbf{DeepHeteroIoT}}              & \textbf{94.38 \%} & \textbf{92.01\%} & \textbf{89.42\%} & \textbf{87.34\%} & \textbf{96.01\%} & \textbf{95.93\%} \\ \hline
\end{tabular}
\end{table}

Additionally, the accuracy scores and macro-average F1-scores provided in Table \ref{tab:comp-with-mace-all} unequivocally demonstrate that DeepHeteroIoT consistently outperforms all alternative combinations involving the previous state-of-the-art machine learning model, MACE \cite{mace_paper}, across all datasets. Our model eventually outperforms MACE \cite{mace_paper} by absolute 2.18 \%, 2.92 \%, 5 \%, in accuracy and 3.31 \%, 3.34 \%, 4.89 \% in macro-average F1-score on the respective Urban Observatory, Swiss Experiment, and IOWA ASOS datasets. It is worth noting that MACE employs an ensemble method alongside a brute-force strategy, which leads to considerable computational costs and undermines its scalability. On the other hand, our proposed deep learning model is a complete end-to-end solution which makes it scalable to the increase of dataset or problem size. For instance, the brute-force approach proposed in the MACE paper \cite{mace_paper} took about 2 hours and 17 minutes to train and build on the smallest dataset (Swiss), while our proposed deep learning model only required approximately 23 minutes and 55 seconds. Moreover, previous studies employing ensemble machine learning techniques \cite{mace_paper,wise_paper} did not incorporate the learning of heterogeneous patterns from IoT sensor data during the training phase. Our proposed deep learning architecture fills this gap by enabling end-to-end learning of complex contextual patterns in challenging IoT sensor data sequences.

\section{Conclusion}
In this study, we proposed a novel deep learning model that integrates learnable local features using CNN and global features using Bi-GRU in one end-to-end network, leading to significantly improved classification of heterogeneous IoT sensor data. Through rigorous experiments across three IoT datasets, we validated that our proposed deep learning model not only outperforms baseline machine learning and deep learning methods but also achieves state-of-the-art performance in classifying heterogeneous IoT sensor data. In the future, we would like to extend our model to multi-modal IoT datasets that occur from a diverse range of mobile computing environments.


\section*{Acknowledgements} This research was funded by the Australian Research Council (ARC) under ARC Discovery Project (DP220101420) and partially supported by the Australian Government Research Training Program (RTP) Allowance Scholarship (Relocation).

\end{document}